%% file: Sparse_Neural_Networks.tex
\documentclass[conference]{IEEEtran}
\pdfoutput=1
\usepackage{cite}
\usepackage{amsmath,amssymb,amsfonts}
\usepackage{algorithmic}
\usepackage{graphicx}
\usepackage{textcomp}
\usepackage{xcolor}
\def\BibTeX{{\rm B\kern-.05em{\sc i\kern-.025em b}\kern-.08em
    T\kern-.1667em\lower.7ex\hbox{E}\kern-.125emX}}
\graphicspath{ {Images/} }

\begin{document}

\title{Training Behavior of Sparse Neural Network Topologies}

\author{\IEEEauthorblockN{Simon Alford$^1$, Ryan Robinett$^1$, Lauren Milechin$^2$, Jeremy Kepner$^{1,3}$
\\
\IEEEauthorblockA{$^1$MIT Mathematics Department, $^2$MIT Department of Earth, Atmospheric, and Planetary Sciences, \\ $^3$MIT Lincoln Laboratory Supercomputing Center}
}}

\maketitle

\begin{abstract}
    Improvements in the performance of deep neural networks have often come
    through the design of larger and more complex networks.  As a result, fast
    memory is a significant limiting factor in our ability to improve network
    performance.  One approach to overcoming this limit is the design of sparse
    neural networks, which can be both very large and efficiently trained.  In
    this paper we experiment training on sparse neural network topologies.  We
    test pruning-based topologies, which are derived from an initially dense
    network whose connections are pruned, as well as RadiX-Nets, a class of
    network topologies with proven connectivity and sparsity properties.
    Results show that sparse networks obtain accuracies comparable to dense
    networks, but extreme levels of sparsity cause instability in training,
    which merits further study.
\end{abstract}


\section{Introduction} \let\thefootnote\relax\footnotetext{
    This material is based in part upon work supported by the NSF under grant
    number DMS-1312831. Any opinions, findings, and conclusions or
    recommendations expressed in this material are those of the authors and do
    not necessarily reflect the views of the National Science Foundation.
}

Neural networks have become immensely popular in recent years due to their
ability to efficiently learn complex representations of data. In particular,
innovations in the design and training of deep convolutional neural networks
have led to remarkable performances in the field of computer vision
\cite{alexnet, vgg, googlenet, resnet}. Researchers have found that making
networks larger, wider, and deeper is a surefire way to increase performance.
Many improvements to network performance come from optimizations which tame the
training hiccups and computational demands which result from using larger,
deeper networks.  Because of the explosion in the size of state-of-the-art
networks, fast memory has become a key limit in our ability to improve neural
network performance.
\par
One strategy to decreasing the memory requirements of training large neural
networks is to introduce sparsity into a network's connections
\cite{kumar2018ibm,KepnerGilbert2011,kepner2017enabling,kepner2018mathematics}.
This strategy has a biological motivation: the human brain exhibits very high
sparsity with its connections, with each neuron connected to approximately 2000
of the 86 billion total neurons on average, a sparsity of $2 \cdot 10^{-8}$
\cite{brain}. In addition, research has shown that trained neural networks
contain many unnecessary weights \cite{predict_params}. If we can discover
sparse topologies which have 'built-in' resolution of these redundancies, then
we could build larger and better networks. Sparse network
structures also lend themselves to being trained in a high-performance manner
via sparse matrix libraries, which would lead the way for extremely
large, sparse neural networks to be efficiently trained. The
MIT/Amazon/IEEE Graph Challenge now features a Sparse Deep Neural Network Graph
Challenge for developing such a system \cite{graph_challenge}.

\par Existing research on sparsity in neural networks is mostly concerned with
model pruning, where large networks are pruned to a small fraction of the original
size without a loss in accuracy. Pruning was introced by Lecun \cite{optimal},
who pruned weights via second derivative information. The work of Han \textit{et
al.} \cite{iterative} introduced the effective technique of iterative pruning, where one alternates training and
pruning a network to increasing levels of sparsity. Several other model
compression techniques have been used, including
low-rank approximation \cite{low_rank}, variational dropout \cite{variational},
and other pruning variations \cite{simple_prune, DNS, explore_sparsity,
to_prune}. 
\par 
While there is a large body of research on model pruning, these methods
typically begin by training a large dense network before pruning to obtain a
sparse network. Such methods aid in model compression for the purpose of model
deployment, but are not applicable to the challenge of desigining sparse
networks for the purpose of efficient \textit{training}. There is little research
addressing the training of purely sparse networks, or the design of originally
sparse network topologies. Even so, sparsity has played an indirect role in
several deep learning innovations. For example, convolutional layers in a
network can be understood as a structured sparsity of connections, allowing more
complex connections to be made without an explosion of
computation requirements. One paper which does address
the creation and training of sparse network topologies is the work of Prabhu
\textit{et al.} \cite{deep_expander}. They replace convolutional and
fully-connected layers with sparse approximations to create sparse networks,
achieving a net decrease in memory required to train the network
without any loss in accuracy.

\par In this paper we continue the search for viable sparse network topologies
for training and large-scale inference. We focus on the trainability of sparse topologies and how
they compare to their dense counterparts. We test two types of sparse
topologies. The first are sparse topologies which result from pruning a dense
network. We prune a pre-trained dense network with both one-time pruning and the
iterative pruning technique developed in \cite{to_prune}. Then we train a new
network with the pruned structure. The second type of sparse topologies we test
are RadiX-Nets. RadiX-Nets, developed by Robinett and Kepner \cite{RadiX-Net},
improve off the work in \cite{deep_expander} to provide sparse topologies with
theoretical guarantees of sparsity and connectivity properties. We replace
fully-connected and convolutional layers with RadiX-Nets and their random
counterparts and train them, comparing the accuracy achieved. Our experiments
are done on the Lenet-5 and Lenet 300-100 models trained on the MNIST
\cite{lenet5} and CIFAR-10 \cite{cifar10} datasets.

\par We find that both the topology of the sparse network and the sparsity level
of the network affect its ability to learn. With pruning-based sparse
topologies, the iteratively-pruned sparse structure could be retrained to higher
accuracy than the one-time pruned structure, giving evidence that specific
connections are crucial to keep when desigining sparse topologies. At higher
sparsity levels, the sparse networks exhibit convergence difficulties. Better
accuracies were obtained on the extremely sparse topologies with a higher learning rate
than that used for training dense networks. With RadiX-Net topologies,
sparser structures perform generally close to the original dense networks, with
50\% sparse networks performing almost equally to their dense
counterparts. The results suggest that for this sparse structure, higher
sparsity levels may limit performance even when the number of total connections is
kept constant.

\section{Training sparse neural networks} Compared to dense networks, there are
many more possibilities for designing the structure of sparse networks,
as we need to decide which subset of a fully-connected layer will be maintained. We
consider two approaches to specifying sparse network structure: pruning-based
structures and Radix-Net structures.

\begin{figure} \begin{center} \includegraphics[width=8cm]{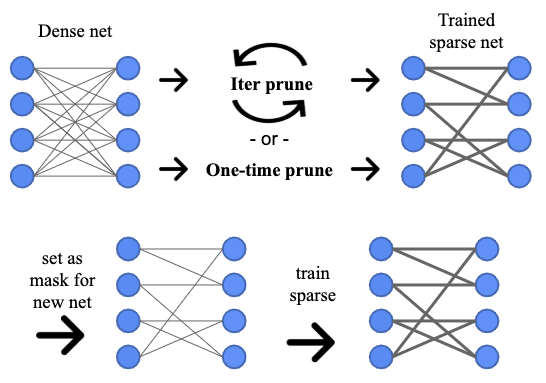}
\end{center} \caption{An overview of the pruning and retraining process.}
\end{figure}

\subsection{Pruning-based structures} \par Our first method uses model pruning
to create a sparse network from a densely-trained network. Among several pruning
techniques, we found the pruning technique from \cite{to_prune} to be most
efficient and effective. The authors made their pruning code open-source, allowing easy
validation and repetition of results. After an initial training period, we prune
the network every 200 steps such that the network sparsity matches a given
sparsity function $s(t)$, a monotonically-increasing function which starts at
zero and finishes at the desired sparsity. Whenever connections are pruned, the
corresponding weights remain zero for the rest of the procedure. After the desired sparsity level is
reached, we train the network for another period with no pruning. This pruned
network is then used as a sparse network structure for a new network, which is
trained with new weights.

\begin{figure}
    \begin{center}
        \includegraphics[width=8cm]{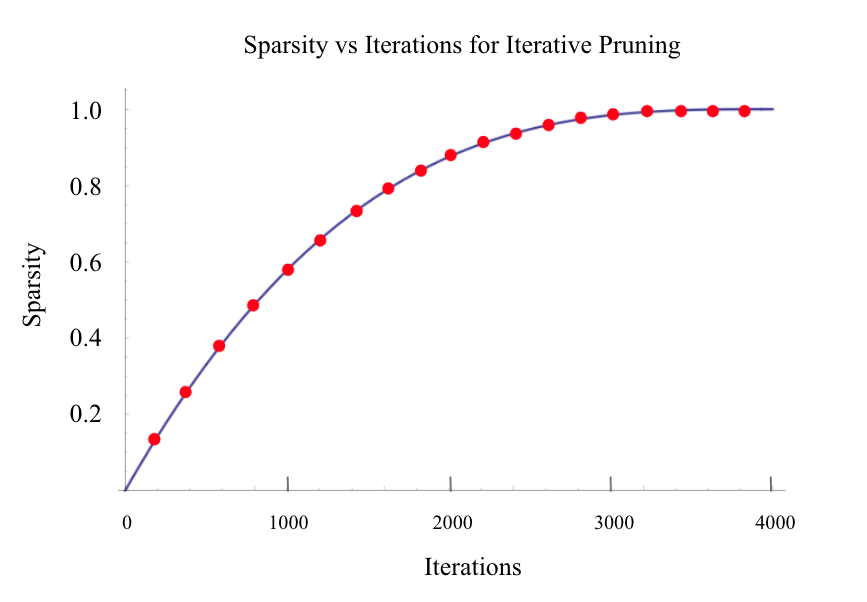}
    \end{center}
    \caption{Pruning is done every 200 steps}
\end{figure}

\par We are able to achieve 99\% sparsity with less than 1\% reduction in accuracy,
and approximately 95\% sparsity without any loss in accuracy whatsoever (before retraining
on the sparse structure). These results show that the sparse structures learned
from the pruning process have the potential to perform just as well as dense
structures. We then see if retraining from scratch on the pruned structure can
recover that accuracy.

\par We prune Lenet-5 and Lenet-300-100 on the MNIST dataset to 50\%, 75\%,
90\%, 95\%, and 99\% sparsity and trained on the sparse structure. To see how
the original accuracy of the sparse model affects the sparsely-trained model's
accuracy, we conduct the same experiment using one-time pruning.  With one-time
pruning, we prune a percentage of the connections with the smallest weights,
without any retraining (leading to lower initial accuracy than the
iteratively-pruned models).

\begin{figure}
    \begin{center}
        \includegraphics[width=8cm]{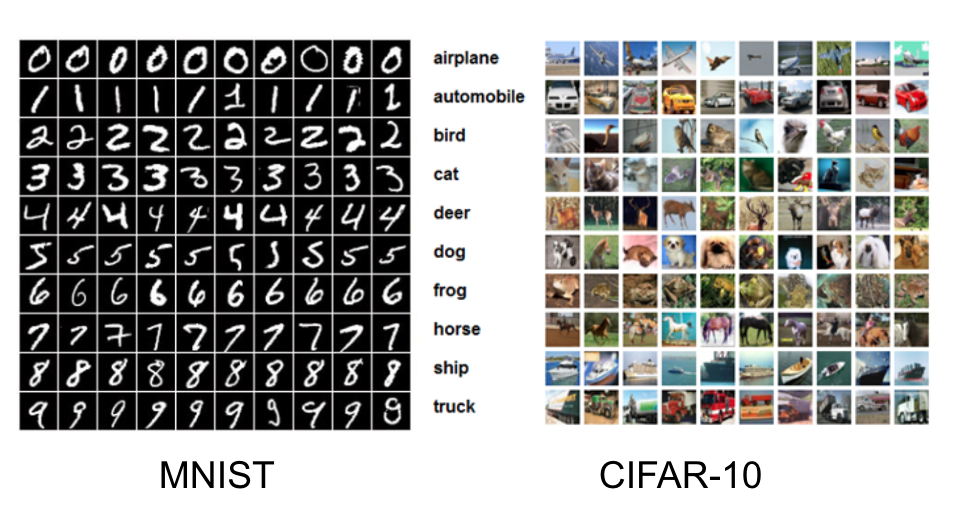}
    \end{center}
    \caption{The MNIST and CIFAR-10 datasets used for experiements.}
\end{figure}

\begin{figure}
    \begin{center}
        \includegraphics[width=8cm]{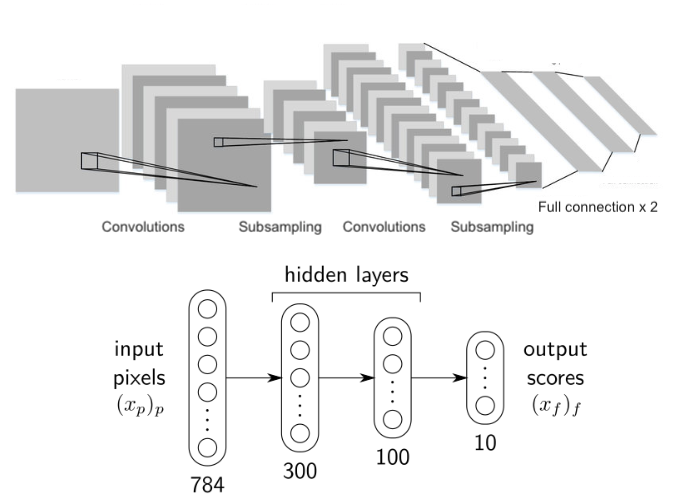}
    \end{center}
    \caption{The Lenet-5 and Lenet 300-100 networks used for experiments.}
\end{figure}

\subsection{RadiX-Net structures} \par Our second method improves on the work done
in \cite{deep_expander}. We replace fully-connected and convolutional layers
with sparse equivalents. We use the RadiX-Nets of Robinett and Kepner
\cite{RadiX-Net} to create our sparse structure. Their work improves off
\cite{deep_expander} by providing more rigorous theoretical guarantees of
path-connectedness between the input and output. A RadiX-Net is defined with a
mixed radix system denoted with a set of $\mathcal{N} = (N_1, N_2, \ldots N_i)$
and a Kronecker structure denoted with a set of $\mathcal{B} = (B_1, B_2, \ldots
B_{i+1})$.

\begin{figure} \begin{center} \includegraphics[width=3cm]{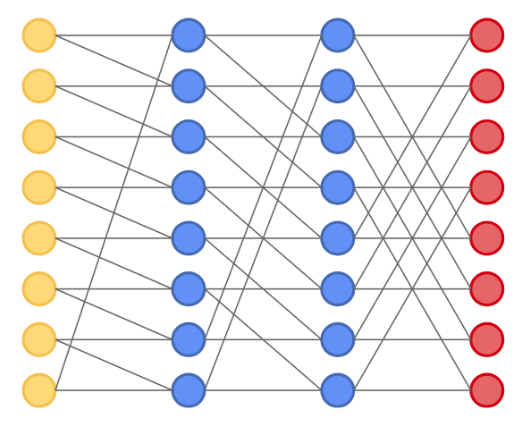}
    \includegraphics[width=3cm]{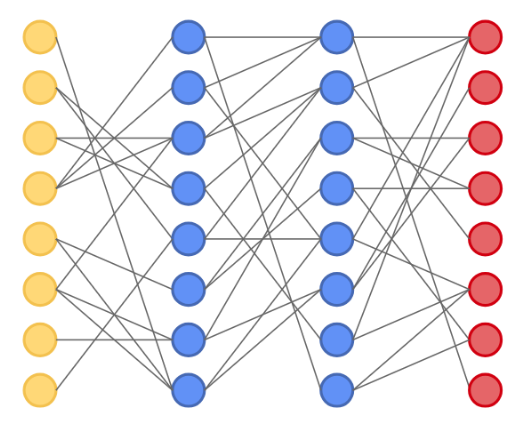} \caption{A two-layer RadiX-net with
    radix values (2, 2, 2) and 75\% sparsity, with its random equivalent
beside.} \end{center} \label{radix} \end{figure}

\par In addition, we test random sparse networks, where given the dense network
to replace, each edge is present with probability $1 - s$, where $s$ is the
desired sparsity level from zero to one. Such layers asymptotically hold the
same theoretical guarantees of path-connectedness as the RadiX-Net while being
simpler to implement. As a result, the majority of our experiments use random
sparse layers. If one were to implement a neural network training framework
using sparse matrix representation, the structured sparsity of RadiX-Nets would
yield much greater computation acceleration compared to the random structure.

\par We experiment with both fixing the number of total connections as sparsity
increases, so that the total neurons increases correspondingly, and fixing the
number of neurons as sparsity increases, so that the total connections decreases
correspondingly. First, we test networks with two, ten, and a hundred times the
number of neurons of the original network, with corresponding sparsities so that
the total connections remain constant. In addition, we test networks with
one-half, one-tenth, one-twentieth, and one-hundredth the number of connections
of the original network, with corresponding sparsities so that the total neurons
remain constant. We train sparse versions of Lenet-300-100 and Lenet-5 on MNIST
as well as Lenet-5 on CIFAR-10. We also test RadiX-Nets on Lenet-300-100 for
networks one-tenth and ten times the size of the original to confirm that their
performance was equal to that of the random sparse nets. Our large and small
Lenet-300-100 RadiX-Nets use $\mathcal{N} = (10, 10), \mathcal{B} = (8,30,1)$
and $\mathcal{N} = (10, 10), \mathcal{B} = (8,3,1)$ respectively to create
networks with the same shape as Lenet-300-100 but with 90\% sparsity. To match
MNIST input dimension of 784, the top 16 neurons are removed from the network.
For more information on the motivation, construction and properties of RadiX-Nets, refer to \cite{RadiX-Net}.

\begin{figure} \begin{center} \includegraphics[width=8cm]{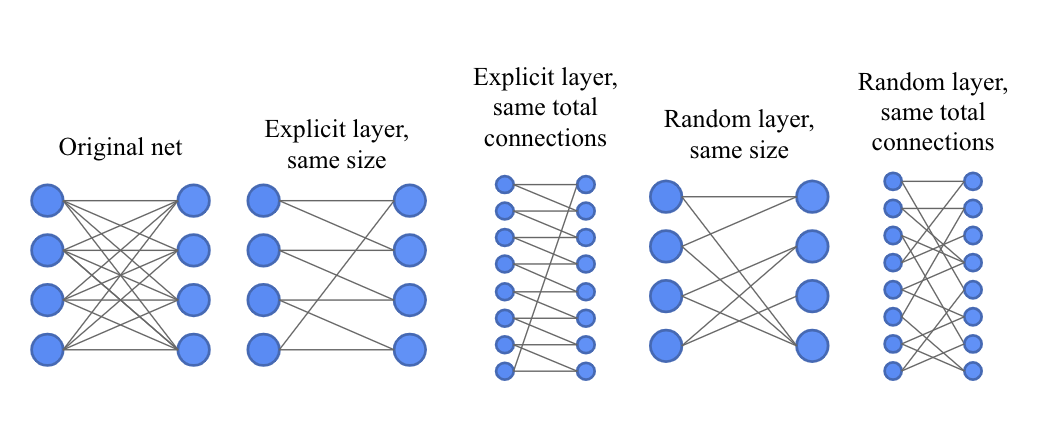}
\end{center} \caption{Examples of the different types of RadiX-Nets tested.}
\label{zoo} \end{figure}

\begin{figure} \begin{center} \includegraphics[width=8cm]{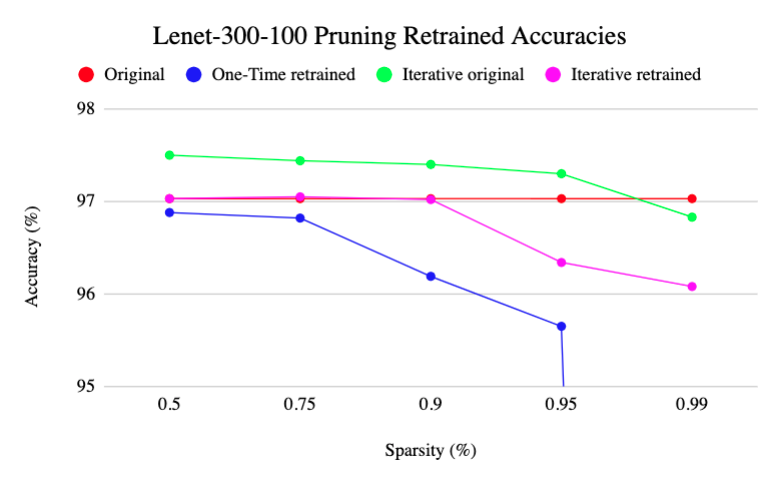}
\includegraphics[width=8cm]{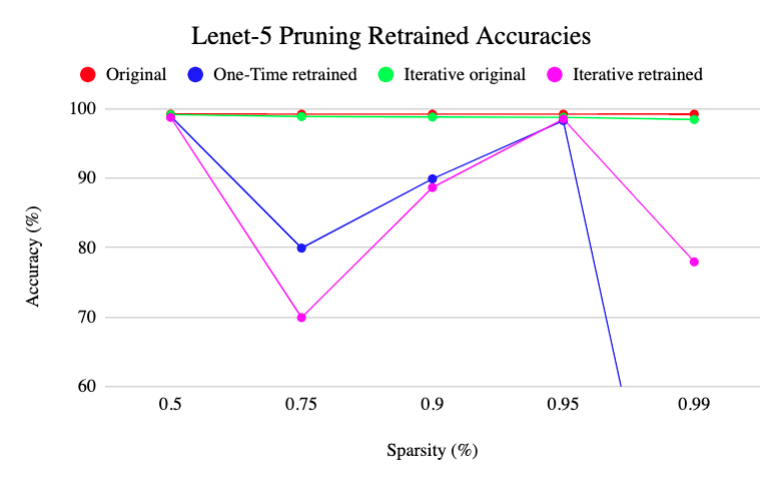} \end{center} \caption{Pruning
retrained accuracies for Lenet-300-100 and Lenet-5} \label{pruning} \end{figure}

\begin{figure*} 
    \centering
        \includegraphics[width=2\columnwidth]{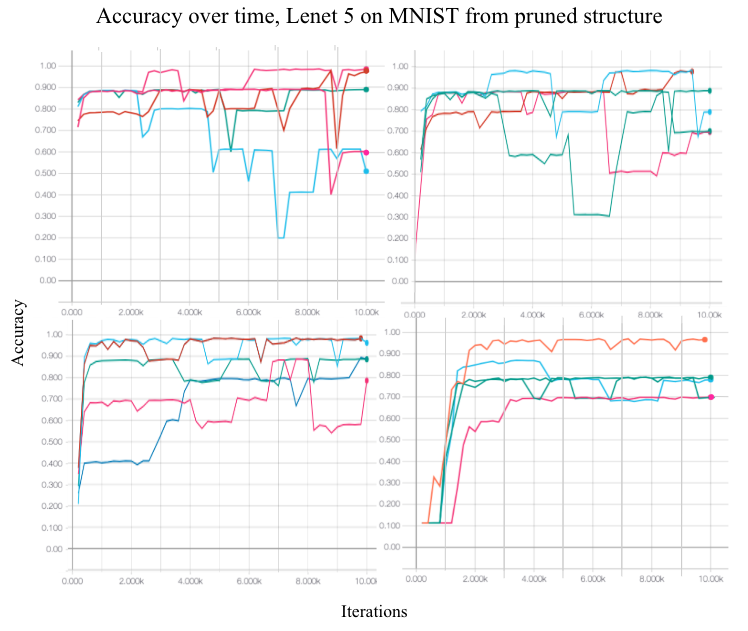}
    \caption{From top left to bottom right, the result of training Lenet-5 on
    MNIST with pruned sparse structures of 0.75, 0.9, 0.95, and 0.99 percent
    sparsity. The figures show the instability of training on the
    sparse pruned network considering multiple Lenet-5 runs. Different colors
    represent retraining on the same sparse structure with everything identical
    except for different weight initializations.} 
    \label{training} 
\end{figure*}

\section{Results} Figure \ref{pruning} shows the result of pruning and
retraining on MNIST. We first see that for Lenet-300-100 through 95\% sparsity,
the iterative pruning actually improves the network's accuracy, with pruning
acting as a form of regularization. Training on the pruned structure does not
reach as high accuracy as the initial accuracy from iterative pruning. However,
up to 90\% it achieves the original accuracy. Training on the structure given by
one-time pruning, however, performs noticeably worse. This suggests that there
are important connections or patterns of connections which one-time pruned
structures do not maintain.

\par The results on Lenet-5 are much different. Unlike Lenet-300-100, this
network contains convolutional layers and achieves much better performance on
MNIST. Even though the iterative pruning process allows the network to obtain
99\% sparsity with less than one percent drop in accuracy, trainability on the
pruned sparse structure exhibits a large variance. After some runs the network
is able to achieve almost the same accuracy, as exhibited by the one-time and
iteratively-pruned models achieving 98.32\% and 98.54\% accuracy for the 95\%
sparse model. However, in general the models fare poorly upon retraining. Figure
\ref{training} below gives more insight into the training process, showing test
accuracy measured throughout the training process for the sparse networks. The
networks appear to get stuck at different levels throughout the process.
Higher learning rates improved this behavior but did not get rid of it
completely. This
behavior suggests that training on sparse networks hinders the ability of
stochastic gradient descent to converge on a solution. While we know from the
pruning process that a sparse solution with high accuracy exists, the
sparsely-trained model may not recover the same accuracy. One can imagine
that having fully-connected layers to train over gives the network flexibility
when searching the space of parameters to optimize for a best representation. As
it is iteratively pruned, the network slowly settles in to a solution. However, when
these pruned connections are set in stone, previously found solutions are now
obstructed by high-cost, suboptimal representations. In order to train sparse
networks, one needs a way to incorporate sparse structure without causing this
sort of roughness in the space of stochastic gradient descent.

\par One factor that may be causing this instability of training is that our
pruning process prunes all layers of the Lenet-5, including the first
convolutional layer. The first convolutional layer is the most important to the
network performance, so pruning it may be limiting the results. However, the
behavior is even seen at relatively low levels of sparsity such as 75\%. In
addition, the accuracy achieved seems uncorrelated with sparsity. More research
into the training process is needed to fully understand why the training process
is less stable for sparse convolutional neural networks.

\begin{figure} \begin{center} \includegraphics[width=8cm]{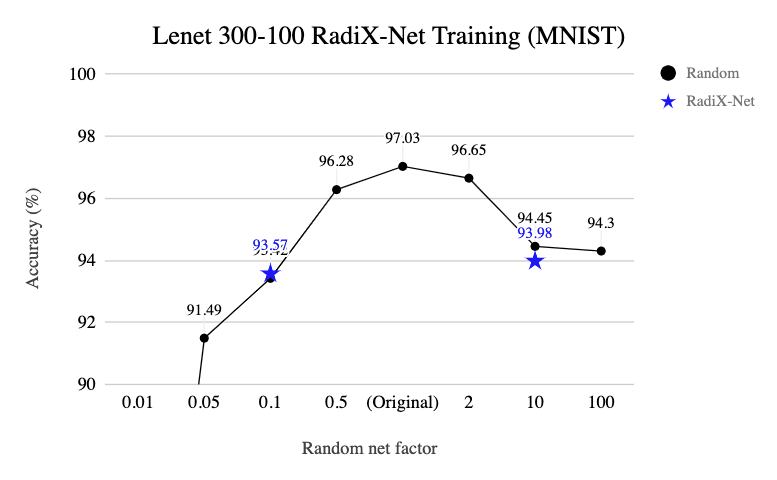}
\includegraphics[width=8cm]{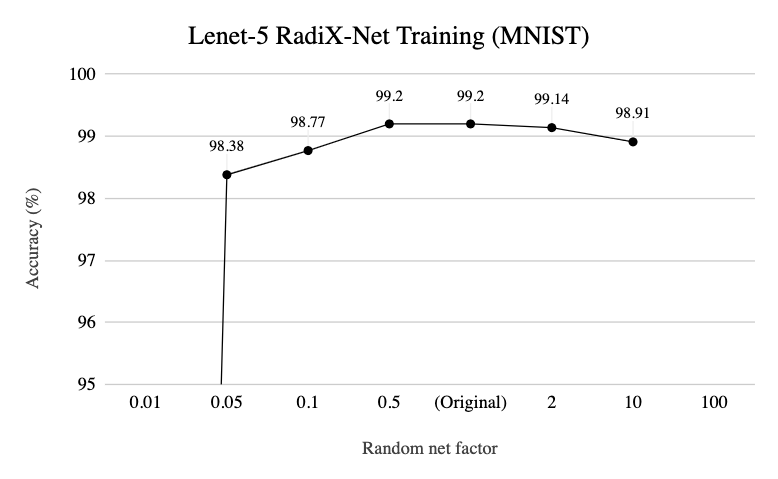}
\includegraphics[width=8cm]{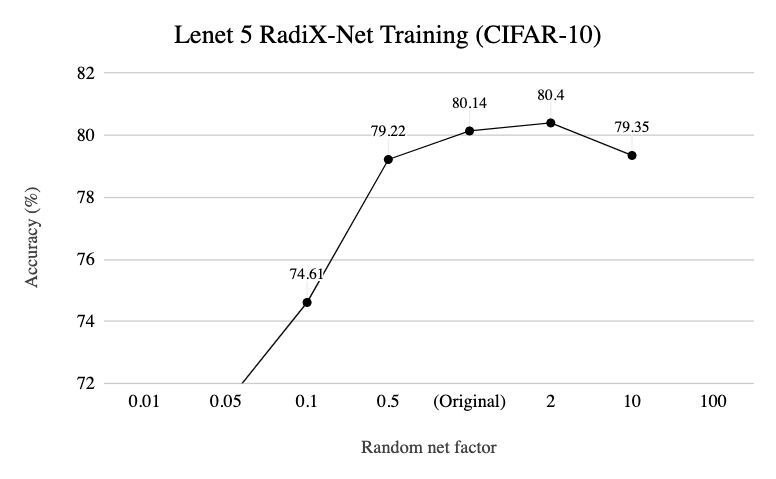} \end{center} \caption{RadiX-Net
accuracies for Lenet-5 and Lenet-300-100} \label{radix} \end{figure}

\par Figure \ref{radix} shows results training the RadiX-Nets on MNIST for
Lenet-300-100 and Lenet-5 and CIFAR-10 for Lenet-5. (Lenet-300-100 is too small
a network to achieve a meaningful result on CIFAR-10.) Lenet-300-100 performs
best at full density, and accuracy decreases as sparsity is increased, even
while keeping total connections constant. We see the same trend for Lenet-5 on
MNIST, but the effect of sparsity is much smaller; it loses only one percent
accuracy through 95\% sparsity. (Training Lenet-5 100 times larger was not
feasible due to memory constraints.) In comparison, the curve for Lenet-5 on
CIFAR-10, a much more challenging dataset, is similar to that for Lenet-300-100.
One explanation for the divergence in behavior for Lenet-5 on MNIST is that
Lenet-5 is overparameterized for the relatively simple MNIST
dataset, and hence can afford the sparsity without being penalized. Still,
sparse versions of Lenet-5 perform very well on both datasets when the total
connections are kept constant. This suggests that if the number of connections is
the only concern, then sparse representations are at least as good as dense
representations.

\par Note that RadiX-Nets leave the first and last layer of Lenet-5 dense. As a
result, any issues Lenet-5 may have had with pruning the first convolutional
layer during the pruning-based sparse training are not present here. More
experiments are needed to see how pruning or preserving certain layers of a
network affects performance.

\section{Conclusion} We trained sparse neural network structures and compared
their performance to their dense counterparts to learn more about how
sparse networks train. Using pruning-based and RadiX-Net sparse network
structures led to different insights in training sparse neural networks. In
general, we found that while sparse networks are able to perform just as well as
dense networks in some cases, increased sparsity can make the training
process less stable.

\par Future work should focus on obtaining the same state-of-the-art accuracies
of dense networks through the use of sparse layers and connectivity topologies.
After the same performance can be reached with sparse networks, the network size
can be increased dramatically to obtain even better results.
\par Another avenue of future work is further investigation into how stochastic
gradient descent behaves when using sparse versus dense network structures.
Understanding how sparsity affects model convergence could be key in designing
sparse structures which train efficiently and effectively.

\par Lastly, in order to fully realize the potential benefits of sparse neural
networks, work needs to be done developing a sparse neural network training
framework which efficiently stores and computes sparse matrices. Traditional
dense models enjoy the enjoy a large body of work optimizing computation on GPUs
specifically for neural network training, and much work will be needed for
sparse matrices to be competitive, even if novel sparse structures showcase
exceptional training potential.

\section*{Acknowledgments}
The authors wish to acknowledge the following
individuals for their contributions and support: Alan Edelman, Vijay Gadepally,
Chris Hill, Richard Wang, and the MIT SuperCloud Team.

\bibliographystyle{ieeetran}
\bibliography{kepner_bib}
\input{Sparse_Neural_Networks.bbl}

\end{document}